# Forget the Learning Rate, Decay Loss

Jiakai Wei

*Abstract*—In the usual deep neural network optimization process, the learning rate is the most important hyper parameter, which greatly affects the final convergence effect. The purpose of learning rate is to control the stepsize and gradually reduce the impact of noise on the network. In this paper, we will use a fixed learning rate with method of decaying loss to control the magnitude of the update. We used Image classification, Semantic segmentation, and GANs to verify this method. Experiments show that the loss decay strategy can greatly improve the performance of the model.

## I. INTRODUCTION

While deep neural networks have achieved amazing successes in a range of applications, how to understand its internal principles is still an open and active area of research. In order to train a model, we need to set a series of hyper parameters, a simple way is to use grid search to find the best value, no doubt it will greatly increase the training time. Maybe we'll choose parameters based on our own empirical, but it's too difficult to beginners, and this is one of the reasons why deep learning is called black box.

For learning rate, a typical training strategy is piecewise constant strategy [1], until [2], [3] proposed that cyclical learning rates can effectively accelerate convergence, after that [4] proposed trapezoid schedule can be further improved. Understanding that adapting the learning rate is a good thing to do, particularly on a per parameter basis dynamically, led to the development of a family of widely-used optimizers including [5], [6]. However, a persisting commonality of these methods is that they are parameterized by a "pesky" fixed global learning rate hyperparameter which still needs tuning.

In this paper, we're going to use an unprecedented way to solve this problem, which we named "Loss Decay". By applying a dynamic weight to the loss (Also called cost, it usually can be derived from the cost function in the model training phase.) can avoid gradient vanishing/exploding. Fig. 1 provides a comparison of test accuracies from a loss decay and normal training regime for Cifar-10, both using a network that is extremely deep and no skip connections. The loss decay strategy experiment uses a linear decrease schedule (see Fig. 2(a)) from 2 to 0, it can be concluded from the experiment that the loss decay strategy can effectively propagate the gradient and improve performance when the model becomes deeper. The contributions of this paper are:

- We propose to use loss decay strategy to adjust the gradient size, it can make the model converge at a fixed learning rate.
- Loss decay strategy can speed up convergence and improve model accuracy, we demonstrate its superiority through image classification, semantic segmentation, and GANs.

## II. BACKGROUND

Since deep learning entered the field of vision in 2012, researchers have been challenging various fields, such as object recognition from images [7]; speech recognition [8]; natural language processing [9], whether it is the network structure [10] or the optimization algorithm [5], [11] has a great improvement.

But we still face the "alchemy" problem, one of which is that the deep neural network has many hyper parameters need to be adjusted. In nearly all gradient descent algorithms the choice of learning rate remains central to efficiency; Reference [12] asserts that it is "often the single most important hyper-parameter and that it always should be tuned." This is because choosing to follow your gradient signal by something other than the right amount, either too much or too little, can be very costly in terms of how fast the overall descent procedure achieves a particular level of objective value.

JK. Wei [13] shows that the "random gradient" method which multiply the gradient by a random number from 0 to 1 can effectively avoid the oscillation of the optimization process. But they did not propose a theoretical explanation, and multiplying the gradient by a random number is too mysterious. This paper gave us the initial inspiration, we will compare this method in subsequent section.

In ordinary training process, we usually use the stochastic gradient descent method or its variants as the optimization algorithm. These methods are widely used in various models, but its shortcomings are also obvious, the learning rate must be shrunk to compensate for even stronger curvature, as a result, learning can become extremely slow. In this paper, we will apply a dynamic loss decay weight to resist the stronger curvature while keeping the learning rate constant. From this we hope we can avoid the problem caused by the increase of $g^T H g$ during training, exploring the relationship between learning rate and gradient, and propose a theoretical analysis.

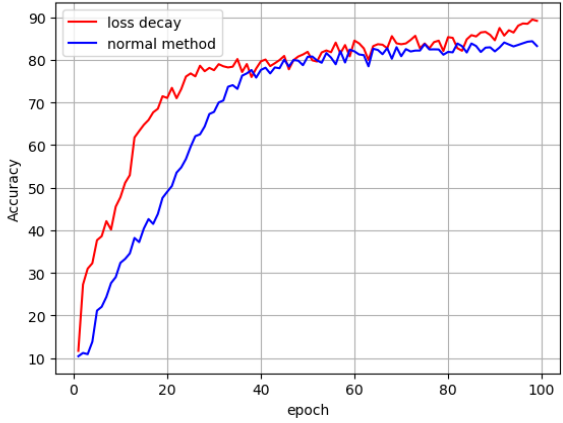 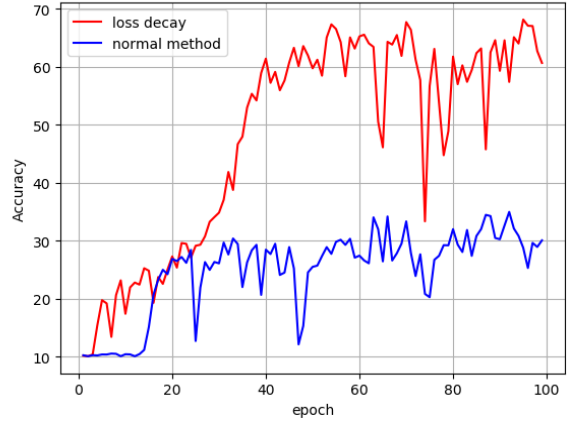

(a): The network is 40 layers and no residual connection.   (b): The network is 50 layers and no residual connection.

Fig. 1: Loss decay represents the method proposed in this paper, it can be seen from the figure that the effect is more obvious when the network depth increases.

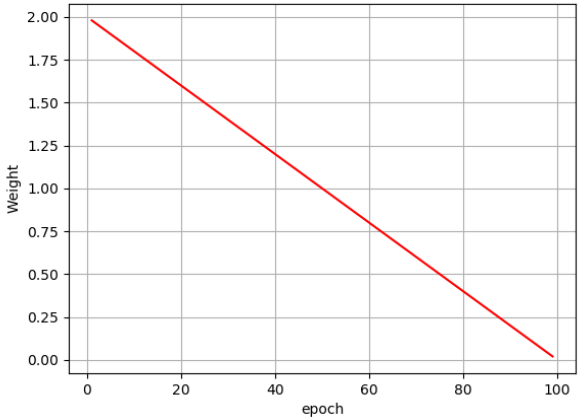 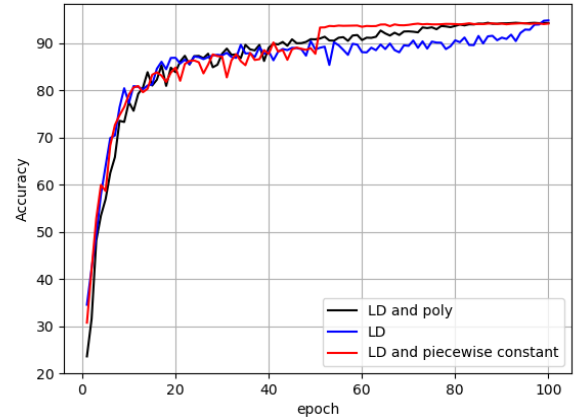

(a): Linear decrease schedule designed for the loss in this paper.

(b): Comparison of gradient weight invariance and linear decrease when fixed poly decay strategy

Fig. 2:" LD" stands for linear decrease strategy," poly" represents polynomial learning rate decay strategy.

### A. Implementation Method

Gradient descent is an optimization method that uses the slope as computed by the derivative to move in the direction of greatest negative gradient to iteratively update a variable. That is, given an initial point $x_0$, gradient descent proposes the next point to be:

$$x = x_0 - \eta \frac{dL}{dx_0} \qquad (1)$$

When η is the learning rate, it can be seen that our method changed the size of $\frac{dL}{dx_0}$. There is an easy way to implement this method, loss (cost) is calculated in most machine learning frameworks [14], [15], changing the weight of loss is equal to changing the weight of the gradient if there are without special training tricks. Since the derivation process will indirectly lead to changes in the gradient, and the purpose of the decay loss is also to adjust the gradient, gradient weights are used instead of loss in the following sections. So we use gradient decay strategy to instead of loss decay for paper's preciseness.

## III. EXPERIMENT AND ANALYSIS

In this section, we will use a lot of space on image classification to explain our method, and then we will verify our conclusions on semantic segmentation and GANs.

### A. Image Classification

As we all know, deep neural networks (DNNs) are a complex and non-convex function, Goodfellow et al. [16] introduced neural network optimization states:

*In the late stage of optimization, learning will become very slow despite the presence of a strong gradient because the learning rate must be shrunk to compensate for even stronger curvature.*

In Fig. 2(b), when using linear decrease strategy (Fig. 2(a)), the model has a very amazing boost at about the last 10 epochs when using a fixed learning rate schedule, we called it "Death Convergence". When we look back equation 1, both the learning rate and gradient determine the next update, but in this paper, we find that the gradient is more effective than the learning rate, because decay learning rate will cause the convergence be slower, but decay gradient will alleviate this problem. Jiakai. Wei [13] explains that applying weights to gradients can slow down the problems caused by hessian matrix with ill-conditioning.

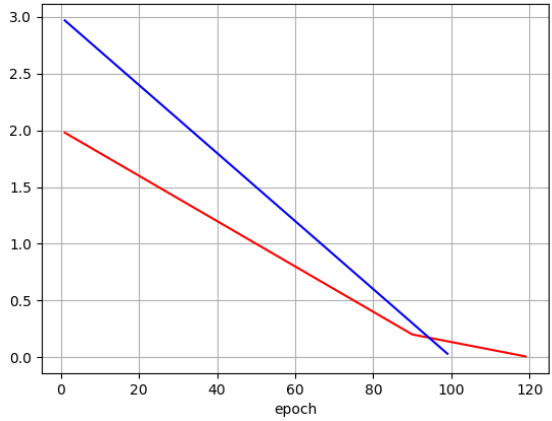

(a): Improved weighting strategy for death convergence.

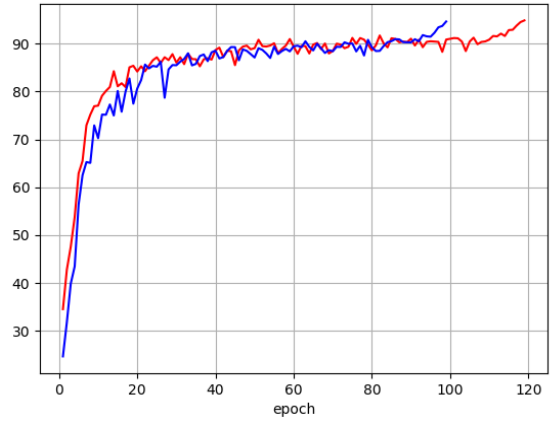

(b): Network results on two improved schedules.

Fig. 3: Further investigation into death convergence.

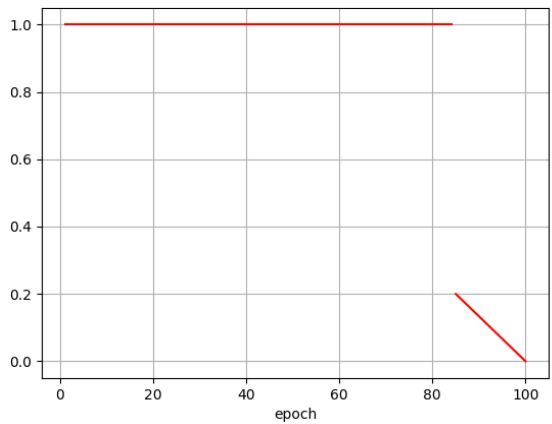

(a): new weighting strategy for death convergence.

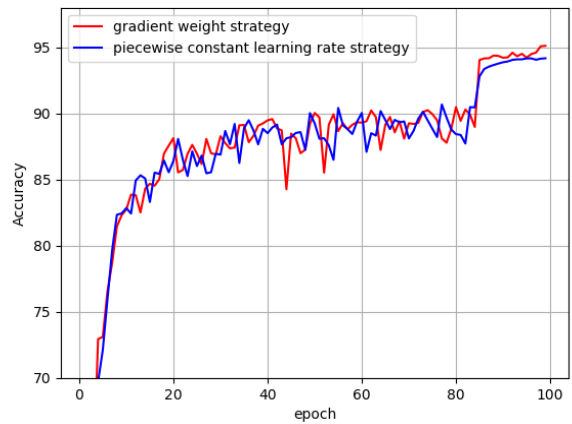

(b): Network results on new weighting strategy.

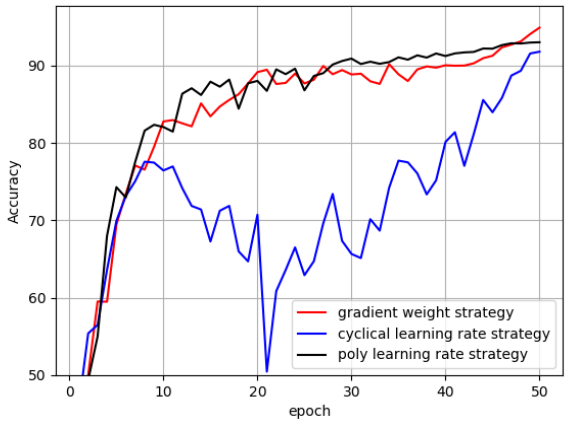

(c): comparison of three different methods.

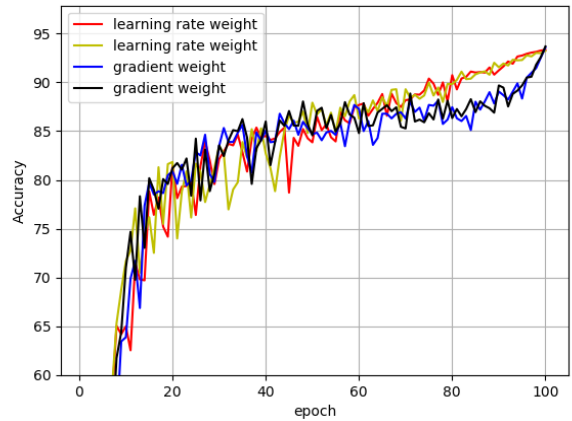

(d): comparing the relationship between gradient weight strategy and learning rate strategy.

Fig. 4: Further investigation into gradient weight strategy.

According to the linear decrease rule in Fig. 2(a), the gradient weights of the last 10 epoch are between 0.2 and 0. For further exploration, we designed a long tail relaxation schedule and linear decrease schedule with a maximum value of 3, as shown in Fig. 3. But it didn't bring any effect, the model still improved rapidly in the last 10 epochs, regardless of the weight of the last 10 epochs. In order for researchers to simply reproduce the phenomenon of death convergence, we will provide the pytorch [17] version code in https://github.com/leemathew1998/GradientWeight.

*1) Theoretical Analysis*

It can be noted in Fig. 3(b) that the phenomenon of death convergence does not occur exactly on the last 10 epochs, but it seems to start inadvertently. The following assumption is crucial to the analysis:

Assumption: We can assume that when the gradient is enough reduced to break the current training deadlock, death convergence can be turned on.

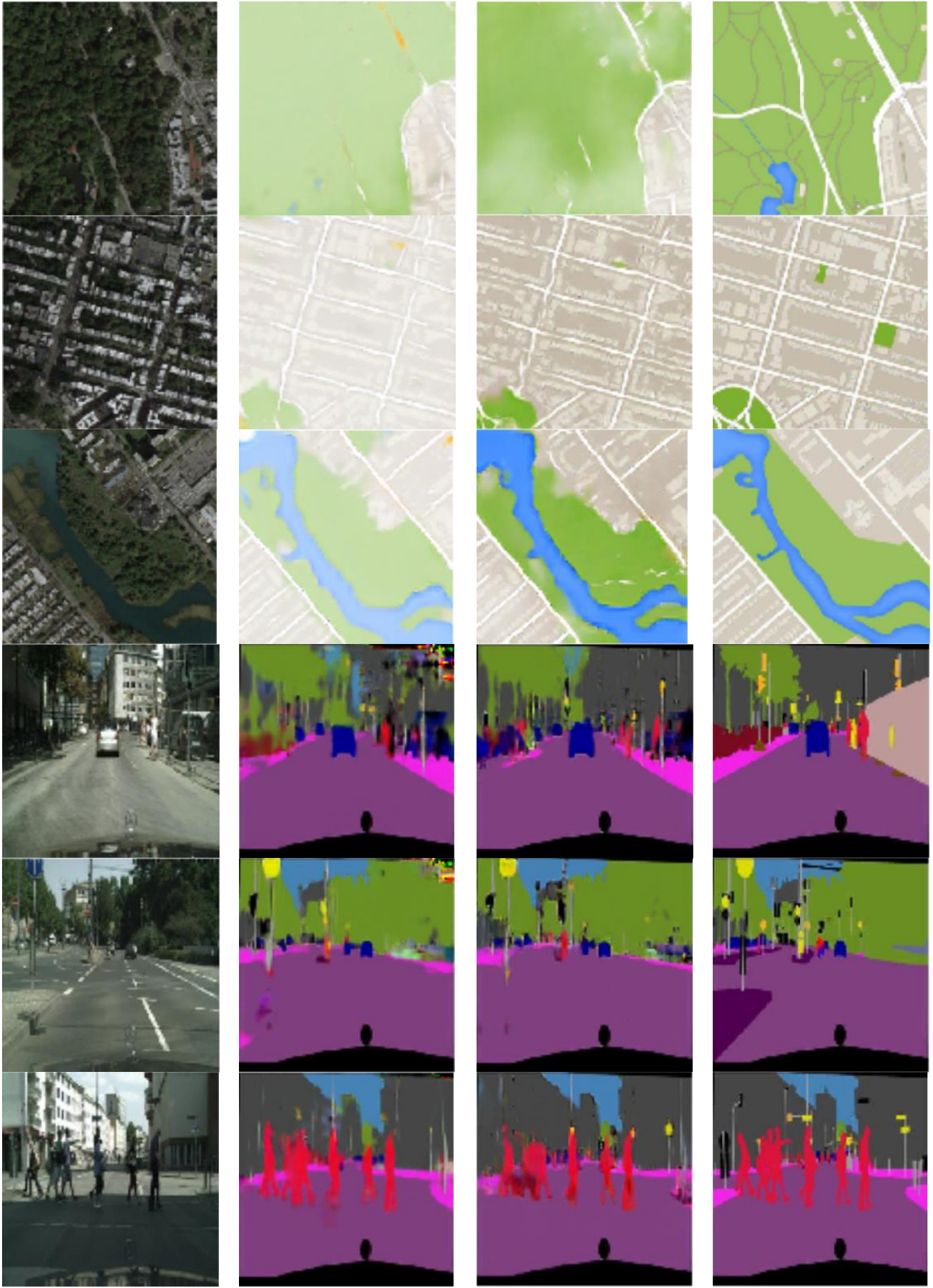

Fig. 6: Visualization result on maps and cityscapes, from left to right are Input, Original Method, Gradient Weight, Ground Truth respectively.

Table 1: This experiment uses PSPNet101 sets the batch size is 4 and the crop size is 224×224, running on a single GTX1060.

| Method | Mean IOU |
|---|---|
| Baseline | 62.13% |
| Gradient Weight | 63.56% |

The most common training method is when the model stops converge is to decay the learning rate, which factor is typically 0.1. When decay two or three times, the model becomes unable to converge through the decay learning rate, normally, this means that training can be stop. In this paper, by using gradient weight strategy, the model can automatically decide the time to start death convergence, this also reduced the burden on researchers. To explain the death convergence, we can start with the second-order Taylor series expansion of the cost function:

$$f(x) \approx f(x_0) - \eta g^T g + \frac{1}{2}\eta^2 g^T H g \qquad (2)$$

Goodfellow et al. [16] states: There are three terms here: the original value of the function, the expected improvement due to the slope of the function, and the correction we must apply to account for the curvature of the function. In many cases, the gradient norm does not shrink significantly throughout learning, but the $g^T H g$ term grows by more than an order of magnitude.

When the model has been oscillating without any performance improvement, it can be considered that $g^T H g$ is already large enough to affect convergence. The result is that learning becomes very slow despite the presence of a strong gradient, and the model will continue to oscillate. For that the gradient weighting strategy has come to the fore, on the premise of no loss of convergence speed, it not only reduces the instability caused by gradient noise, but also makes the model break the current deadlock and converge further.

This raises a problem: Whether this method can be approximated to adjust the learning rate? The answer is negative. In Fig. 4(d), we apply the linear decrease schedule, which used on the loss, to the learning rate, it can be seen that there is a significant difference between them, and it is also verified that the gradient weight strategy does not improve by indirectly adjusting the learning rate.

Or a more direct explanation: In the normal training process, we only change the learning rate, and the value of the weight decay is a constant, but in this paper, we change the weight so that the gradient gradually decreases while the value of the weight decay is constantly changing. This little change has also triggered our thinking. Why do we need a constant value for weight decay instead of a changed value? Why weight decay does not change with the learning rate? In the future, these issues can be studied in depth.

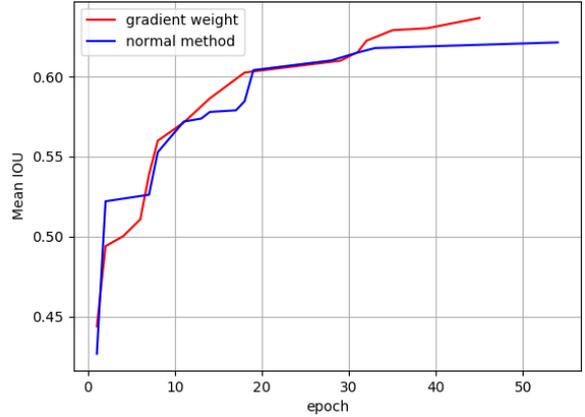

Fig. 5: Model performance when used PSPNet for test.

*1) Comparison of Results*

Based on the above analysis, we have designed a new weighting strategy in Fig. 4(a), Fig. 4(b) shows a comparison of this method with the piecewise constant learning rate schedule. From the figure we can see that the model with the gradient weight strategy can converge better, the final result is 1% better than the piecewise constant learning rate schedule. Fig. 4(c) shows the convergence of the model we specified in 50 epochs, we can see that the death convergence played a decisive role in the final stage, leading the second place by nearly 2%.

*B. Semantic Segmentation*

Semantic segmentation with the goal to assign semantic labels to every pixel in an image [18] is one of the fundamental topics in computer vision. Deep convolutional neural networks [19] based on the Fully Convolutional Neural Network [20] show striking improvement over systems relying on hand-crafted features [21, 22, 23] on benchmark tasks.

This experiment contains 20 foreground object classes and one background class, dataset [18] contains 1,464 (train), 1,449 (val), and 1,456 (test) pixel-level annotated images. The performance is measured in terms of pixel intersection-over-union averaged across the 21 classes (mIOU), but the commonly used extra annotations datasets [24] will not be used to improve accuracy. Inspired by Hariharan et al. [25], we use the "poly" learning rate policy that the current learning rate equals to the base one multiplying $(1 - \frac{iter}{max\_iter})^{power}$. We set the base power to 0.9, we use the random mirror for data augmentation. Inspired by Szegedy et al. [18], we set the momentum to 0.95, learning rate to 0.001, and experiment with the PSPNet [26]. From Table 1 and Fig. 6, it is concluded that the gradient weight strategy can converge faster and better.

But it is very intuitive to feel that the Mean-IOU of this article does not meet the highest standards, the main reason is that all the experiments in this article are run on a GTX1060 graphics card. In the future, we will test in a more powerful GPU.

*C. GAN*

Generative Adversarial Networks (GANs) [27, 28] have achieved impressive results in image generation [29], and representation learning [30]. The key to GANs' success is the idea of an adversarial loss that forces the generated images to be, in principle, indistinguishable from real images. This is particularly powerful for image generation tasks, as this is exactly the objective that much of computer graphics aims to optimize. GANs learn a loss that tries to classify if the output image is real or fake, while simultaneously training a generative model to minimize this loss. Blurry images will not be tolerated since they look obviously fake. Because GANs learn a loss that adapts to the data, they can be applied to a multitude of tasks that traditionally would require very different kinds of loss functions.

We used the excellent pix2pix [31] network to experiment. In Fig. 5, it can be seen clearly that our method can generate clearer and more realistic images than usual, especially on the map dataset, the color produced by the gradient weight strategy is more realistic. In all the experiments in this paper, GAN is a very intuitive demonstration of the improvements brought by the gradient weight strategy.

In generating tasks, we accord the method mentioned in the paper to do our experiments. We apply the Adam solver [32], with learning rate 0.0002, and momentum parameters $\beta 1 = 0.9$, $\beta 2 = 0.999$, we trained the network for 200 epochs, please refer to the original paper for details.

## IV. CONCLUSION

The results in this paper presented the benefits of the gradient weight strategy, apply a weight less than 1 to the gradient at the end of the training, usually have 1 to 2 percent improvement in the field of image classification, 1 percent improvement in the field of semantic segmentation, generate clearer and more realistic images in GAN, prove the feasibility of gradient weight strategy in the field of computer vision. And found that this method unique death convergence phenomenon can converge faster and better, the experimental results prove that it is more effective than adjusting the learning rate.

## V. LIMITATION AND FUTURE WORK

The most obvious drawback is that we don't have enough machines to get the model to the top performance, but the conclusions we have made as far as possible from a fair experiment are still convincing. Another drawback is even more obvious, that is, we did not present a convincing mathematical explanation, just the explanation based on the experimental results does not satisfy us.

This article is far from over, not to mention that the above two drawbacks can be improved, we can also design a new gradient weight strategy for model training. And it's not in all experiments that this strategy can lead to improvement, for example, in GAN experiment, there are many models that produce similar or difficult results to determine which model is better, but this is also a disadvantage of generative adversarial networks itself. In addition, all of the above experiments belong to the field of computer vision, and we are not sure how it works in other fields.

## VI. ACKNOWLEDGEMENT

Even though there are too many places to improve, but I still want to thank my classmate Pengsheng Xu for modifying the mistakes in this article. He was extremely patient and kind, without his help, I will not be able to complete this article, when everyone is not optimistic, support me as always. Thus, I would like to wish him a brilliant career life.

I hope this paper will not be the end of my academic thinking, although I am still very interested in the whole machine learning field, the reality is very cruel. If I still insist on academic research after five years, I will thank myself for being able to withstand the pressure. If not, I will respect every decision I make.

## REFERENCES

[1] Yann Le Cun. Efficient backprop. Neural Networks Tricks of the Trade, 1524(1):9–50, 1998.

[2] Leslie N. Smith. Cyclical learning rates for training neural networks, 2015. arXiv:1506.01186.

[3] Leslie N. Smith and Nicholay Topin. Super-convergence: very fast training of residual networks using large learning rates., 2017.

[4] Chen Xing, Devansh Arpit, Christos Tsirigotis, and Y Bengio. A walk with sgd. 02 2018.

[5] T. Tieleman and G. Hinton. Coursera: Neural network for machine learning, 2012. Lecture 6.5 - RMSProp.

[6] J. Duchi, E. Hazan, and Y. Singer. Adaptive subgradient methods for online learning and stochastic optimization. Journal of Machine Learning Research, 12:2121–2159, 2011.

[7] Kaiming He, Xiangyu Zhang, Shaoqing Ren, and Jian Sun. Deep residual learning for image recognition. pages 770–778, 2016. In CVPR.

[8] Dario Amodei, Rishita Anubhai, Eric Battenberg, Carl Case, Jared Casper, Bryan Catanzaro, Jingdong Chen, Mike Chrzanowski, Adam Coates, and Greg Diamos. Deep speech 2: End-to-end speech recognition in english and mandarin. Computer Science, 2015.

[9] Minh Thang Luong, Hieu Pham, and Christopher D. Manning. Effective approaches to attention-based neural machine translation. Computer Science, 2015.

[10] Christian Szegedy, Vincent Vanhoucke, Sergey Ioffe, Jon Shlens, and Zbigniew Wojna. Rethinking the inception architecture for computer vision. In IEEE Conference on Computer Vision and Pattern Recognition, pages 2818–2826, 2016b.

[11] Geoffrey E. Hinton. A practical guide to training restricted boltzmann machines. Momentum, 9(1):599–619, 2012.

[12] Y. Bengio. Practical recommendations for gradient-based training of deep architectures. In Neural Networks: Tricks of the Trade, volume 7700, pp. 437–478. Springer, 2012.

[13] JK. Wei. Faster, Better Training Trick --- Random Gradient. (unpublished). arXiv:1808.04293.

[14] Martın Abadi, Ashish Agarwal, Paul Barham, Eugene Brevdo, Zhifeng Chen, Craig Citro, Greg S. Corrado, Andy Davis, Jeffrey Dean, and Matthieu Devin. Tensorflow: Large-scale machine learning on heterogeneous distributed systems. 2015. Yann Le Cun. Efficient backprop. Neural Networks Tricks of the Trade, 1524(1):9–50, 1998.

[15] Tianqi Chen, Mu Li, Yutian Li, Min Lin, Naiyan Wang, Minjie Wang, Tianjun Xiao, Bing Xu, Chiyuan Zhang, and Zheng Zhang. Mxnet: A flexible and efficient machine learning library for heterogeneous distributed systems, 2017. arXiv:1512.01274.

[16] Ian Goodfellow, Yoshua Bengio, and Aaron Courville. Deep Learning. MIT Press, 2016. http://www.deeplearningbook.org, pp. 123-157.

[17] Adam Paszke, Sam Gross, Soumith Chintala, Gregory Chanan, Edward Yang, Zachary DeVito, Zeming Lin, Alban Desmaison, Luca Antiga, and Adam Lerer. Automatic differentiation in pytorch, 2017.

[18] Mark Everingham, Luc Van Gool, Christopher K. I. Williams, John Winn, and Andrew Zisserman. The pascal visual object classes (voc) challenge. International Journal of Computer Vision, 88(2):303–338, 2010.

[19] LeCun, Y., Bottou, L., Bengio, Y., Haffner, P.: Gradient-based learning applied to document recognition. In: Proc. IEEE. (1998)

[20] Sermanet, P., Eigen, D., Zhang, X., Mathieu, M., Fergus, R., LeCun, Y.: Overfeat: Integrated recognition, localization and detection using convolutional networks. In: ICLR. (2014)
[21] He, X., Zemel, R.S., Carreira-Perpindn, M.: Multiscale conditional random fields for image labeling. In: CVPR. (2004)
[22] Simonyan, K., Zisserman, A.: Very deep convolutional networks for large-scale image recognition. In: ICLR. (2015)
[23] Szegedy, C., Liu, W., Jia, Y., Sermanet, P., Reed, S., Anguelov, D., Erhan, D., Vanhoucke, V., Rabinovich, A.: Going deeper with convolutions. In: CVPR. (2015)
[24] Marius Cordts, Mohamed Omran, Sebastian Ramos, Timo Rehfeld, Markus Enzweiler, Rodrigo Benenson, Uwe Franke, Stefan Roth, and Bernt Schiele. The cityscapes dataset for semantic urban scene understanding. In IEEE Conference on Computer Vision and Pattern Recognition, pages 3213–3223, 2016.
[25] B. Hariharan, P. Arbelaez, L. Bourdev, S. Maji, and J. Malik. Semantic contours from inverse detectors. In ICCV, 2011.
[26] H. Zhao, J. Shi, X. Qi, X. Wang, and J. Jia. Pyramid scene parsing network. In CVPR, 2017.
[27] Ian J. Goodfellow, Jean Pouget-Abadie, Mehdi Mirza, Bing Xu, David Warde-Farley, Sherjil Ozair, Aaron Courville, and Yoshua Bengio. Generative adversarial nets. In International Conference on Neural Information Processing Systems, pages 2672–2680, 2014.
[28] J. Zhao, M. Mathieu, and Y. LeCun. Energy-based generative adversarial network. In ICLR, 2017.
[29] Emily Denton, Soumith Chintala, Arthur Szlam, and Rob Fergus. Deep generative image models using a laplacian pyramid of adversarial networks. pages 1486–1494, 2015.
[30] Tim Salimans, Ian Goodfellow, Wojciech Zaremba, Vicki Cheung, Alec Radford, and Xi Chen. Improved techniques for training gans. 2016.
[31] Phillip Isola Jun-Yan Zhu Tinghui Zhou Alexei A. Efros. Image-to-Image Translation with Conditional Adversarial Networks. Journal of Machine Learning Research, 2016.
[32] D. Kingma and J. Ba. Adam: A method for stochastic optimization. ICLR, 2015.

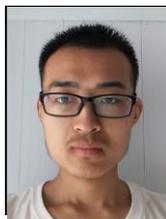

**Jiakai Wei** was born in Shan Dong (China) in 1998. He is currently a student at the Hunan University of Technology, because he is an undergraduate student, and has not achieved any outstanding results. His main research interests are optimization algorithms.